\documentclass{article}

\usepackage[final]{neurips_2019}

\usepackage[utf8]{inputenc} 
\usepackage[T1]{fontenc}    
\usepackage{hyperref}       
\usepackage{url}            
\usepackage{booktabs}       
\usepackage{amsfonts}       
\usepackage{nicefrac}       
\usepackage{microtype}      

\usepackage{tabularx}
\usepackage{amsmath,array,graphicx}
\usepackage{multirow}
\usepackage{ctable}
\usepackage{hhline}
\usepackage[toc,page]{appendix}
\usepackage{booktabs,subcaption,amsfonts,dcolumn}
\usepackage{titlesec}

\usepackage{natbib}

\newcommand{\vect}[1]{\boldsymbol{#1}}
\newcommand{\vadv}{\mathit{v\textnormal{-}adv}}

\titlespacing*{\section}
{0pt}{0.8ex plus 0.7ex minus .2ex}{0.6ex plus .2ex}
\titlespacing*{\subsection}
{0pt}{0.8ex plus 0.7ex minus .2ex}{0.6ex plus .2ex}
\titlespacing*{\subsubsection}
{0pt}{0.8ex plus 0.7ex minus .2ex}{0.6ex plus .2ex}

\title{Advancing PICO Element Detection in Biomedical Text via Deep Neural Networks}

\author{%
  Di Jin, Peter Szolovits \\
  Compter Science and Artificial Intelligence Laboratory\\
  Massachusetts Institute of Technology \\
  Cambridge, MA, 02139, USA \\
  \texttt{{jindi15,psz}@mit.edu} \\
}

\begin{document}

\maketitle

\begin{abstract}
  In evidence-based medicine (EBM), defining a clinical question in terms of the specific patient problem aids the  physicians to efficiently identify appropriate  resources  and  search  for  the  best  available  evidence  for medical treatment. In order to formulate a well-defined, focused clinical question, a framework called PICO is widely used, which identifies the sentences in a given medical text that belong to the four components typically reported in clinical trials: Participants/Problem (P), Intervention (I), Comparison (C) and Outcome (O). In this work, we propose a novel deep learning model for recognizing PICO elements in  biomedical  abstracts.  Based  on  the  previous  state-of-the-art  bidirectional long-short term memory (biLSTM) plus conditional random field (CRF) architecture,   we  add  another  layer  of  biLSTM  upon  the  sentence representation vectors so that the contextual information from surrounding sentences can be gathered to help infer the interpretation of the current one. In addition, we propose two  methods  to  further  generalize  and improve the  model:  adversarial  training  and unsupervised pre-training over large corpora. 
 We tested our proposed approach over two benchmark datasets. One is the PubMed-PICO dataset, where our best results outperform the previous best by 5.5\%, 7.9\%, and 5.8\% for P, I, and O elements in terms of F1 score, respectively. And for the other dataset named NICTA-PIBOSO, the improvements for P/I/O elements are 2.4\%, 13.6\%, and 1.0\% in F1 score, respectively. Overall, our proposed deep learning model can obtain unprecedented PICO element detection accuracy while avoiding the need for any manual feature selection. 
\end{abstract}

\section{Introduction}

In evidence-based medicine (EBM), well formulated and structured documents and questions can help physicians efficiently identify appropriate resources and search for the best available evidence for medical treatment \citep{richardson1995well}. In practice, clinical studies and questions always either explicitly or implicitly contain four aspects: Population/Problem (P), Intervention (I), Comparison (C) and Outcome (O), which are known as PICO elements. Using this structure to help with the information retrieval (IR) of medical evidence within a large medical citation database is popular and advantageous \citep{huang2006evaluation, schardt2007utilization,
boudin2010improving}.
However, accurately and efficiently extracting PICO elements from non-structured information such as a collection of medical abstracts is challenging.

The PICO element detection process can be cast as a classification task on the sentence or segment level. Previously there have been many studies that sought to develop  algorithms for this problem.  In earlier work, these studies used older machine learning techniques such as Naive Bayes (NB) \citep{huang2013pico}, Random Forest (RF) \citep{boudin2010combining}, Support Vector Machine (SVM) \citep{hansen2008method}, Conditional Random Field (CRF) \citep{kim2011automatic}, and Multi-Layer Perceptron (MLP) \citep{huang2011classification,hassanzadeh2014identifying,chabou2018combination}. All these methods heavily rely on careful collections of features including lexical features such as bag of words (BOW), synonyms and hypernyms, semantic features such as part-of-speech (POS) and named entities, and sequential features such as the relative position of each token. More recently, emerging deep Artificial Neural Network (ANN) architectures such as the bi-directional Long Short-Term Memory (bi-LSTM) model have been adopted to further improve performance \citep{jin2018pico}. That model proposed to encode each sentence into a high-dimensiomal representation vector via  a recurrent neural network (RNN) and then added a CRF module to optimize the label sequence. We call this the ``bi-LSTM+CRF'' architecture. 

Although the preliminary application of deep learning models has demonstrated superior performance compared with shallow machine learning models, there still remain abundant opportunities to enlarge this gap. In this work, we propose to significantly boost the PICO element detection accuracy for deep learning models by exploiting two observations. On one hand, we deem the PICO detection process as a sequential sentence classification problem, where structured predictions need to be made for each sentence in the text. In this scenario, the contextual information from surrounding sentences can be utilized to help infer the label of the current one. Therefore, based on the state-of-the-art (SOTA) ``bi-LSTM+CRF'' architecture, we stack another layer of bi-LSTM over the encoded sentence representation vectors to aggregate the features of surrounding sentences so that the output hidden state vectors carry not only the information of the current sentence but also that of adjacent sentences. 

On the other hand, deep learning models are prone to over-fit over data of small size, resulting in unsatisfactory PICO element extraction performance compared with shallow machine learning models when training data is small. To remedy this issue, we adopted two strategies to enhance the generalization capability of our proposed model. One is to use adversarial and virtual adversarial training to regularize the model by stabilizing the classification function \citep{miyato2016adversarial}. The other is to first pre-train a language model using a large-scale bio-medical literature corpus and then fine-tune on the targeted datasets, which is generally called inductive transfer learning. 
Both strategies can further boost the detection performance significantly.

With all these contributions, we are able to validate the superiority of deep learning models by advancing the PICO element detection task to new SOTA performance by evaluation on two benchmark datasets. Specifically, for one dataset PubMed-PICO, the absolute improvement of F1 score for the three P/I/O elements are 5.5\%, 7.9\%, and 5.8\% in terms of F1 score, respectively. And for the other one named NICTA-PIBOSO, the improvements for P/I/O elements are 2.4\%, 13.6\%, and 1.0\% in F1 score, respectively. We hope such success of our proposed model can encourage the wider incorporation of deep learning models into the PICO framework in Evidence-Based Practice (EBP).

\section{Methods}

\subsection{Model Architecture}


The model architecture is illustrated in Figure~\ref{figure:model}(A). It mainly consists of three modules: the sentence encoder, the sentence contextualization layer, and the label sequence optimization layer. Detailed functions of each module are described below:

\begin{figure*}[!tpb]
\centering
\includegraphics[width=0.9\textwidth]{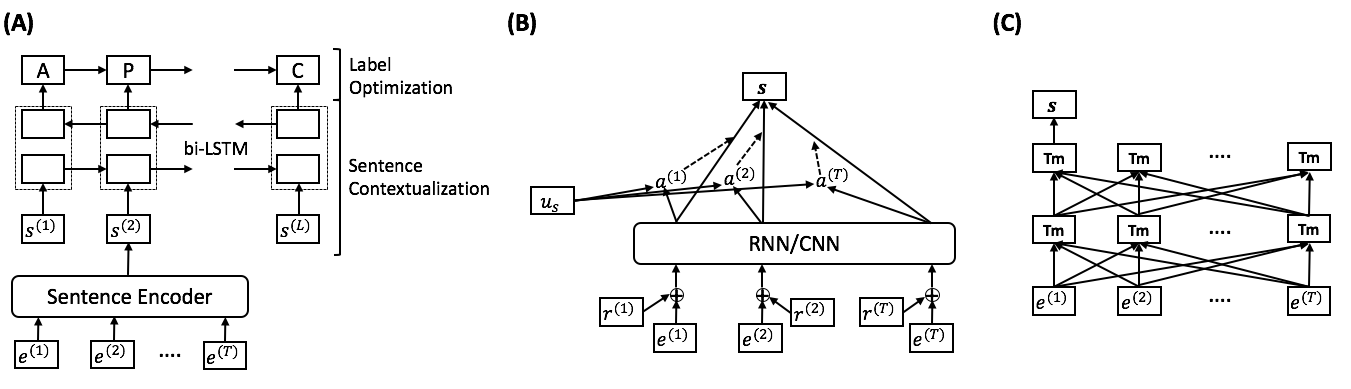}
\caption{Model architecture. (A) Overall model architecture; (B) RNN/CNN based sentence encoder; (C) Transformer based sentence encoder.}
\vspace{-5mm}
\label{figure:model}
\end{figure*}

\paragraph{Sentence Encoder}

This layer takes as input a sequence of tokens and produces a vector $\vect{s}$ to encode this sequence. We proposed two ways to implement this encoder and they will be described in the next subsection.

\paragraph{Sentence Contextualization}

So far, for a paragraph or document consisting of a sequence of sentences, we have obtained a sequence of vectors $\vect{h'}_{1:L}$ ($\vect{h'}\in \mathbb{R}^{d^{h_d}}$), each of which corresponds to a sentence. In this step, these vectors are further processed by a bi-LSTM layer, so that we can contextualize each sentence vector with the information from surrounding sentences. Each of these contextualized vectors is subsequently input to a feed-forward neural network with only one hidden layer to get the corresponding probability vector $\vect{r}\in \mathbb{R}^{l}$, which represents the probability that this sentence belongs to each label, where $l$ is the number of labels.

\paragraph{Label Sequence Optimization Layer}

We finally use a CRF module to optimize the sequence of labels \citep{collobert2011natural}. It can model the dependencies between subsequent labels so that some unlikely label sequences can be avoided.






\subsection{Sentence Encoders}

We propose two kinds of sentence encoders, each of which has its pros and cons. The first one is based on CNN or RNN models and is trained from random initialization. The other is based on the Transformer architecture \citep{vaswani2017attention} and it is first pre-trained over a large text corpus via the language modeling task and then fine-tuned to the target datasets. Overall, the latter encoder outperforms the former one but it contains many more parameters and takes a much longer time for the pre-training step. 
Details of these two encoders are elaborated in the Appendix \ref{appendix:sentence-encoder}.


\section{Experiments}

\subsection{Datasets}

Two datasets were used in this work, described below, whose statistics are summarized in Table \ref{table:data-stats}.  

\paragraph{\textbf{PubMed-PICO}}

This dataset is the benchmark dataset from \citep{jin2018pico}.\footnote{https://github.com/jind11/PubMed-PICO-Detection} One typical data example is shown in Table \ref{table:abstract-example}.

\paragraph{\textbf{NICTA-PIBOSO}}

This corpus was extracted by the authors of \citep{kim2011automatic}.\footnote{https://github.com/jind11/NICTA-PIBOSO-Dataset} 

\subsection{Experimental Settings}

Ten-fold cross-validation was used to report the final performance results, where we divided the full dataset into 10 folds and iteratively used one fold as the development set, one as the test set and the rest as the train set. We report the results using the standard class-based (or micro) precision, recall and F1 scores. The test set was always evaluated at the highest development set performance. This enables us to provide a clear view of the behaviour of the classifier in each class, in addition to comparing our results to prior approaches. For NICTA-PIBOSO dataset, we use the complete ``6-way'' PIBOSO scheme.


The key hyperparameters for the model architecture with the RNN/CNN based sentence encoder are tuned on the development set and summarized in Table \ref{table:hyperparameters} . Since the dataset sizes of the two dataset differ a lot, we obtained different optimal hyperparameter settings for them. Especially we found RNN works better than CNN for the PubMed-PICO dataset while CNN performs better for the NICTA-PIBOSO dataset.

\section{Results}

\subsection{PubMed-PICO Dataset}

Table \ref{table:performance} summarizes the performance results of our proposed model for the PubMed-PICO dataset by comparing with previous results. In this table, the previously published methods for comparison include LR, MLP, CRF, and BiLSTM+CRF, which are all from \cite{jin2018pico}. For our proposed model, there are several variants: the baseline is our proposed architecture based on the RNN/CNN sentence encoder as illustrated in Figure \ref{figure:model} without either adversarial training or virtual adversarial training; ``Adv.'' and ``V-Adv.'' mean that, based on the same architecture, we further use adversarial training or virtual adversarial training when training the model, respectively; ``Adv.+V-Adv.'' means that we use both adversarial training methods, and the unlabeled data for virtual adversarial training are all coming from the original training data; since the virtual adversarial training does not require the data to be labeled, based on the same architecture plus both adversarial training methods, we experimented on adding more unlabeled data by randomly extracting some unannotated abstracts from the PubMed website (we have made sure they do no overlap with the original dataset), and the three numbers ``20,000'', ``100,000'', and ``200,000'' mean the number of unannotated abstracts we used for virtual adversarial training; finally ``'BioBERT'' means that we change the sentence encoder to BERT and pre-train it over the bio-medical literature corpora. 

As we can see from Table \ref{table:performance}, our baseline model improves by a large margin compared with all previous methods for all three P/I/O elements. Especially for the I element, which performs the worst among the three labels, the absolute increase in F1 score is the highest, reaching 5\%. The major change between the baseline model and the previous ``BiLSTM+CRF'' architecture is the addition of the sentence contextualization layer, which indicates that the contextual information extracted by the newly added upper layer of bi-LSTM from surrounding sentences is very helpful for the PICO element extraction, especially for the I element. 

Furthermore, when we adopt adversarial training while optimizing the baseline model, the absolute increases in F1 score for all three P/I/O elements are around 1\%, which demonstrates the effectiveness of adversarial training as a means of regularization. On the other hand, the improvement brought by virtual adversarial training is not as much as adversarial training, which could be because the loss of virtual adversarial training is calculated in an unsupervised way and thus is not specific to this task. However, when we combine the adversarial and virtual adversarial training, we can achieve a larger improvement than using either alone, indicating that these two techniques can complement each other. And with both adversarial training strategies, the absolute improvements of F1 score for the three P/I/O labels are 4.3\%, 6.8\%, and 5.0\%, respectively. 

\begin{table}[!htbp]
\centering
\tiny
\vspace{-3mm}
\caption{Performance of the PubMed-PICO dataset in terms of precision (p), recall (r) and F1 on the test set (average value based on 10-fold cross validation). Best results over all models are marked in bold font and best results over the RNN/CNN based models are marked with underlines.} 
\label{table:performance}
{\begin{tabular}{lccccccccc}
\specialrule{.1em}{.05em}{.05em}
\multirow{2}{*}{\textbf{Models}} & \multicolumn{3}{c}{\textbf{P-element (\%)}} & \multicolumn{3}{c}{\textbf{I-element (\%)}} & \multicolumn{3}{c}{\textbf{O-element (\%)}} \\ \cline{2-10} 
                        & \textbf{p}        & \textbf{r}        & \textbf{F1}       & \textbf{p}        & \textbf{r}        & \textbf{F1}       & \textbf{p}        & \textbf{r}        & \textbf{F1}       \\ \hline
\textbf{LR}                      & 66.9     & 68.5     & 67.7     & 55.6     & 55.0     & 55.3     & 65.4     & 67.0     & 66.2     \\ 
\textbf{MLP}                     & 77.8     & 74.1     & 75.8     & 64.3     & 65.9     & 64.9     & 73.8     & 77.9     & 75.8     \\ 
\textbf{CRF}                     & 82.2     & 77.5     & 79.8     & 67.8     & 70.3     & 69.0     & 76.0     & 76.3     & 76.2     \\ 
\textbf{BiLSTM+CRF}               & 87.8     & 83.4     & 85.5     & 72.7     & 81.3     & 76.7     & 81.1     & 85.3     & 83.1     \\ \hline
\textbf{Ours--Baseline} & 90.0 & 86.6 & 88.3 & 79.6 & 84.0 & 81.7 & 85.5 & 87.8 & 86.6 \\ 
\textbf{Ours--Adv.} & 90.5 & 88.0 & 89.2 & 81.8 & 84.3 & 83.0 & 85.8 & 89.7& 87.7\\
\textbf{Ours--V-Adv.} & 90.2 & 87.8 & 89.0 & 80.7 & 83.3 & 81.9 & 86.3 & 88.6 & 87.4\\ 
\textbf{Ours--Adv.+V-Adv.} & 91.7 & \underline{88.1} & 89.8 & 82.4 & 84.6 & 83.5 & \underline{87.0} & 89.4 & 88.1 \\ \\
\textbf{Ours--20,000} & \underline{92.2} & 87.7 & 89.9 & 82.1 & 84.8 & 83.4 & 86.4 & 89.9 & 88.0\\
\textbf{Ours--100,000} & 92.0 & \underline{88.1} & \underline{90.0} & \underline{83.8} & 83.5 & 83.6 & 86.8 & 89.9 & 88.3\\
\textbf{Ours--200,000} & 92.0 & 87.8 & 89.9 & 83.0 & \textbf{85.1} & \underline{84.0} & 86.7 & \textbf{90.0} & \underline{88.4} \\
\hline
\textbf{Ours--BioBERT} & \textbf{92.8} & \textbf{89.2} & \textbf{91.0} & \textbf{84.1} & 85.0 & \textbf{84.6} & \textbf{88.0} & 89.8 & \textbf{88.9} \\
\specialrule{.1em}{.05em}{.05em}
\end{tabular}}{}
\end{table}

\subsection{NICTA-PIBOSO Dataset}

For the NICTA-PIBOSO dataset, when comparing with previous methods, we found that some of them tested their methods over the official split test set while others used 10 fold cross-validation. Therefore, for fair and comprehensive comparison, we tested our model over both data split schemes and reported the results in Table \ref{table:performance-nicta-official} and Table \ref{table:performancenicta-nicta-10folds}, respectively. To be noted, since this dataset is small, there exists high variance for the results of different runs, therefore we report the average results after five independent runs for Table \ref{table:performance-nicta-official}. By glancing at both tables, we see that overall our proposed models outperform previous works significantly, especially for ``population'', ``intervention'', ``background'', and ``study design'' labels. For example, for the ``intervention'' label, in the official split, our best F1 score exceeds previous SOTA result by 27.8\% in absolute value, while in the cross-validation, the gap between our highest F1 score and the previous highest is 13.6\%. Such significant improvements clearly contradict the meme that deep learning models are not good for small datasets and even perform worse than some simple and shallow machine learning models in such cases. To be mentioned, our models never used the section headings that are provided in the structured abstracts of the dataset as features while most previous works did so. Actually, such section heading information does boost the detection performance significantly due to the high correlation between ground truths and section headings \citep{kim2011automatic,verbeke2012statistical}. Our models do not rely on such information that is only available in structured abstracts but still achieve SOTA performance, which highlights the superiority of our methods.



\begin{table}[b]
\vspace{-3mm}
\caption{Performance of the NICTA-PIBOSO dataset.}
\begin{subtable}[t]{.48\linewidth}
\centering
\caption{Performance of the NICTA-PIBOSO dataset for official train/test split in terms of F1 on the test set. P, I, O, B, SD represent population, intervention, outcome, background, study design labels, respectively. 
} 
\label{table:performance-nicta-official}
\resizebox{0.98\columnwidth}{!}{\begin{tabular}{lcccccc}
\hline
\textbf{Model}             & \textbf{P (\%)}    & \textbf{I (\%)}    & \textbf{O (\%)}   & \textbf{B (\%)}   & \textbf{SD (\%)}  & \textbf{Other (\%)} \\ \hline
\textbf{Marco Lui \citep{lui2012feature}} & 58.0 & 34.0 & 89.0 & 80.0 & 59.0 & \textbf{85.0}  \\
\textbf{A\_MQ \citep{amini2012overview}}  & 51.0 & 35.0 & 86.0 & 78.0 & 58.0 & 84.0  \\ 
\textbf{\cite{dernoncourt2016neural}} & 59.2 & 36.5 & 89.1 & 80.3 & 62.1 & 60.2 \\\hline
\textbf{Ours--Baseline}    &  65.2  & 55.3 & 89.0 &  82.1& 77.8 & 62.2  \\ 
\textbf{Ours--Adv.}        & 66.1&56.8&90.0&  84.0  &  81.4  &   66.1    \\ 
\textbf{Ours--V-Adv.}      & 65.8&   48.5  &  89.3    & 82.6  &  80.9 & 63.5 \\ 
\textbf{Ours--Adv.+V-Adv.} &  \underline{67.7} & \underline{57.6} &  \underline{90.4} & \underline{84.1} & 82.2&  66.3 \\ \\
\textbf{Ours--20,000}      &  66.9 &  50.2  &  \underline{90.4}  & 83.9 & \textbf{82.4} & 67.1 \\ 
\textbf{Ours--100,000}     &  66.3 & 52.6 &  \underline{90.4} & 83.8 & 81.6  &  \underline{67.6}     \\ 
\textbf{Ours--200,000}     &  66.7 &  55.5 &   90.1 &  83.3&  78.2  & 67.5   \\ \hline
\textbf{Ours--BioBERT}     &  \textbf{74.6} & \textbf{64.3} & \textbf{90.9} & \textbf{85.0} & 80.0 & \underline{72.7}\\ \hline
\end{tabular}}
\end{subtable}
\qquad
\begin{subtable}[t]{.48\linewidth}
\centering
\caption{Performance of the NICTA-PIBOSO dataset for 10 fold cross-validation in terms of F1 on the test set. P, I, O, B, SD represent population, intervention, outcome, background, study design labels, respectively. 
}
\label{table:performancenicta-nicta-10folds}
\resizebox{0.99\columnwidth}{!}{\begin{tabular}{lcccccc}
\hline
\textbf{Model}             & \textbf{P (\%)}    & \textbf{I (\%)}    & \textbf{O (\%)}   & \textbf{B (\%)}   & \textbf{SD (\%)}  & \textbf{Other (\%)} \\ \hline
\textbf{\cite{kim2011automatic}} & 47.4 & 16.2 & 81.9 & 74.6 & 22.3 & 45.3  \\
\textbf{\cite{verbeke2012statistical}}  & 28.0 & 20.7 & 84.7 & 81.2 & 24.5 & 53.7  \\ 
\textbf{\cite{sarker2013approach}} & 52.6 & 34.7 & 85.8 & 80.2 & 56.8 & 66.5 \\
\textbf{\cite{chabou2018combination}}
& 73.0 & 43.0 & 90.0 & -- & -- &--\\\hline
\textbf{Ours--Baseline}    &  66.1  & 50.9 & 89.4 &  81.4 & 71.6 & 66.5  \\ 
\textbf{Ours--Adv.}        & 68.1&51.2&90.5&  83.8  &  \textbf{79.2}  &   67.1    \\ 
\textbf{Ours--V-Adv.}      & 66.2 &   47.7  &  89.7    & 82.1  &  72.8 & 67.6 \\ 
\textbf{Ours--Adv.+V-Adv.} &  69.8 & \underline{53.2} &  90.2 & 84.0 & 72.9 &  67.0 \\ \\
\textbf{Ours--20,000}      &  68.7 &  51.7  &  90.6  & \textbf{84.2} & 74.2 & 67.9 \\ 
\textbf{Ours--100,000}     &  69.6 & 52.0 &  \underline{90.8} & 83.4 & 73.4  &  \underline{68.9}     \\ 
\textbf{Ours--200,000}     &  \underline{70.3} &  50.9 & 90.1 &  83.4&  71.5  & 68.3   \\ \hline
\textbf{Ours--BioBERT}    &  \textbf{75.4} & \textbf{56.6} & \textbf{91.0} & 83.9 & 77.1 & \textbf{72.1}\\ \hline
\end{tabular}}
\end{subtable}
\end{table}

\small
\bibliographystyle{unsrtnat}
\bibliography{references}

\begin{thebibliography}{28}
\providecommand{\natexlab}[1]{#1}
\providecommand{\url}[1]{\texttt{#1}}
\expandafter\ifx\csname urlstyle\endcsname\relax
  \providecommand{\doi}[1]{doi: #1}\else
  \providecommand{\doi}{doi: \begingroup \urlstyle{rm}\Url}\fi

\bibitem[Richardson et~al.(1995)Richardson, Wilson, Nishikawa, and
  Hayward]{richardson1995well}
W~Scott Richardson, Mark~C Wilson, Jim Nishikawa, and Robert~SA Hayward.
\newblock The well-built clinical question: a key to evidence-based decisions.
\newblock \emph{ACP journal club}, 123\penalty0 (3):\penalty0 A12--A12, 1995.

\bibitem[Huang et~al.(2006)Huang, Lin, and Demner-Fushman]{huang2006evaluation}
Xiaoli Huang, Jimmy Lin, and Dina Demner-Fushman.
\newblock Evaluation of pico as a knowledge representation for clinical
  questions.
\newblock In \emph{AMIA annual symposium proceedings}, volume 2006, page 359.
  American Medical Informatics Association, 2006.

\bibitem[Schardt et~al.(2007)Schardt, Adams, Owens, Keitz, and
  Fontelo]{schardt2007utilization}
Connie Schardt, Martha~B Adams, Thomas Owens, Sheri Keitz, and Paul Fontelo.
\newblock Utilization of the pico framework to improve searching pubmed for
  clinical questions.
\newblock \emph{BMC medical informatics and decision making}, 7\penalty0
  (1):\penalty0 16, 2007.

\bibitem[Boudin et~al.(2010{\natexlab{a}})Boudin, Shi, and
  Nie]{boudin2010improving}
Florian Boudin, Lixin Shi, and Jian-Yun Nie.
\newblock Improving medical information retrieval with pico element detection.
\newblock In \emph{European Conference on Information Retrieval}, pages 50--61.
  Springer, 2010{\natexlab{a}}.

\bibitem[Huang et~al.(2013)Huang, Chiang, Xiao, Liao, Liu, and
  Wong]{huang2013pico}
Ke-Chun Huang, I-Jen Chiang, Furen Xiao, Chun-Chih Liao, Charles Chih-Ho Liu,
  and Jau-Min Wong.
\newblock Pico element detection in medical text without metadata: Are first
  sentences enough?
\newblock \emph{Journal of biomedical informatics}, 46\penalty0 (5):\penalty0
  940--946, 2013.

\bibitem[Boudin et~al.(2010{\natexlab{b}})Boudin, Nie, Bartlett, Grad, Pluye,
  and Dawes]{boudin2010combining}
Florian Boudin, Jian-Yun Nie, Joan~C Bartlett, Roland Grad, Pierre Pluye, and
  Martin Dawes.
\newblock Combining classifiers for robust pico element detection.
\newblock \emph{BMC medical informatics and decision making}, 10\penalty0
  (1):\penalty0 29, 2010{\natexlab{b}}.

\bibitem[Hansen et~al.(2008)Hansen, Rasmussen, and Chung]{hansen2008method}
Marie~J Hansen, Nana~{\O} Rasmussen, and Grace Chung.
\newblock A method of extracting the number of trial participants from
  abstracts describing randomized controlled trials.
\newblock \emph{Journal of Telemedicine and Telecare}, 14\penalty0
  (7):\penalty0 354--358, 2008.

\bibitem[Kim et~al.(2011)Kim, Martinez, Cavedon, and Yencken]{kim2011automatic}
Su~Nam Kim, David Martinez, Lawrence Cavedon, and Lars Yencken.
\newblock Automatic classification of sentences to support evidence based
  medicine.
\newblock In \emph{BMC bioinformatics}, volume~12, page~S5. BioMed Central,
  2011.

\bibitem[Huang et~al.(2011)Huang, Liu, Yang, Xiao, Wong, Liao, and
  Chiang]{huang2011classification}
Ke-Chun Huang, Charles Chih-Ho Liu, Shung-Shiang Yang, Furen Xiao, Jau-Min
  Wong, Chun-Chih Liao, and I-Jen Chiang.
\newblock Classification of pico elements by text features systematically
  extracted from pubmed abstracts.
\newblock In \emph{Granular Computing (GrC), 2011 IEEE International Conference
  on}, pages 279--283. IEEE, 2011.

\bibitem[Hassanzadeh et~al.(2014)Hassanzadeh, Groza, and
  Hunter]{hassanzadeh2014identifying}
Hamed Hassanzadeh, Tudor Groza, and Jane Hunter.
\newblock Identifying scientific artefacts in biomedical literature: The
  evidence based medicine use case.
\newblock \emph{Journal of biomedical informatics}, 49:\penalty0 159--170,
  2014.

\bibitem[Chabou and Iglewski(2018)]{chabou2018combination}
Samir Chabou and Michal Iglewski.
\newblock Combination of conditional random field with a rule based method in
  the extraction of pico elements.
\newblock \emph{BMC medical informatics and decision making}, 18\penalty0
  (1):\penalty0 128, 2018.

\bibitem[Jin and Szolovits(2018)]{jin2018pico}
Di~Jin and Peter Szolovits.
\newblock Pico element detection in medical text via long short-term memory
  neural networks.
\newblock In \emph{Proceedings of the BioNLP 2018 workshop}, pages 67--75,
  2018.

\bibitem[Miyato et~al.(2016)Miyato, Dai, and Goodfellow]{miyato2016adversarial}
Takeru Miyato, Andrew~M Dai, and Ian Goodfellow.
\newblock Adversarial training methods for semi-supervised text classification.
\newblock \emph{arXiv preprint arXiv:1605.07725}, 2016.

\bibitem[Collobert et~al.(2011)Collobert, Weston, Bottou, Karlen, Kavukcuoglu,
  and Kuksa]{collobert2011natural}
Ronan Collobert, Jason Weston, L{\'e}on Bottou, Michael Karlen, Koray
  Kavukcuoglu, and Pavel Kuksa.
\newblock Natural language processing (almost) from scratch.
\newblock \emph{Journal of Machine Learning Research}, 12\penalty0
  (Aug):\penalty0 2493--2537, 2011.

\bibitem[Vaswani et~al.(2017)Vaswani, Shazeer, Parmar, Uszkoreit, Jones, Gomez,
  Kaiser, and Polosukhin]{vaswani2017attention}
Ashish Vaswani, Noam Shazeer, Niki Parmar, Jakob Uszkoreit, Llion Jones,
  Aidan~N Gomez, {\L}ukasz Kaiser, and Illia Polosukhin.
\newblock Attention is all you need.
\newblock In \emph{Advances in Neural Information Processing Systems}, pages
  5998--6008, 2017.

\bibitem[Verbeke et~al.(2012)Verbeke, Van~Asch, Morante, Frasconi, Daelemans,
  and De~Raedt]{verbeke2012statistical}
Mathias Verbeke, Vincent Van~Asch, Roser Morante, Paolo Frasconi, Walter
  Daelemans, and Luc De~Raedt.
\newblock A statistical relational learning approach to identifying evidence
  based medicine categories.
\newblock In \emph{Proceedings of the 2012 Joint Conference on Empirical
  Methods in Natural Language Processing and Computational Natural Language
  Learning}, pages 579--589. Association for Computational Linguistics, 2012.

\bibitem[Lui(2012)]{lui2012feature}
Marco Lui.
\newblock Feature stacking for sentence classification in evidence-based
  medicine.
\newblock In \emph{Proceedings of the Australasian Language Technology
  Association Workshop 2012}, pages 134--138, 2012.

\bibitem[Amini et~al.(2012)Amini, Martinez, Molla, et~al.]{amini2012overview}
Iman Amini, David Martinez, Diego Molla, et~al.
\newblock Overview of the alta 2012 shared task.
\newblock 2012.

\bibitem[Dernoncourt et~al.(2016)Dernoncourt, Lee, and
  Szolovits]{dernoncourt2016neural}
Franck Dernoncourt, Ji~Young Lee, and Peter Szolovits.
\newblock Neural networks for joint sentence classification in medical paper
  abstracts.
\newblock \emph{arXiv preprint arXiv:1612.05251}, 2016.

\bibitem[Sarker et~al.(2013)Sarker, Moll{\'a}-Aliod, Paris,
  et~al.]{sarker2013approach}
Abeed Sarker, Diego Moll{\'a}-Aliod, Cecile Paris, et~al.
\newblock An approach for automatic multi-label classification of medical
  sentences.
\newblock 2013.

\bibitem[Mikolov et~al.(2013)Mikolov, Sutskever, Chen, Corrado, and
  Dean]{mikolov2013distributed}
Tomas Mikolov, Ilya Sutskever, Kai Chen, Greg~S Corrado, and Jeff Dean.
\newblock Distributed representations of words and phrases and their
  compositionality.
\newblock In \emph{Advances in neural information processing systems}, pages
  3111--3119, 2013.

\bibitem[Pennington et~al.(2014)Pennington, Socher, and
  Manning]{pennington2014glove}
Jeffrey Pennington, Richard Socher, and Christopher Manning.
\newblock Glove: Global vectors for word representation.
\newblock In \emph{Proceedings of the 2014 conference on empirical methods in
  natural language processing (EMNLP)}, pages 1532--1543, 2014.

\bibitem[Bojanowski et~al.(2016)Bojanowski, Grave, Joulin, and
  Mikolov]{bojanowski2016enriching}
Piotr Bojanowski, Edouard Grave, Armand Joulin, and Tomas Mikolov.
\newblock Enriching word vectors with subword information.
\newblock \emph{arXiv preprint arXiv:1607.04606}, 2016.

\bibitem[Kim(2014)]{kim2014convolutional}
Yoon Kim.
\newblock Convolutional neural networks for sentence classification.
\newblock \emph{arXiv preprint arXiv:1408.5882}, 2014.

\bibitem[Yang et~al.(2016)Yang, Yang, Dyer, He, Smola, and
  Hovy]{yang2016hierarchical}
Zichao Yang, Diyi Yang, Chris Dyer, Xiaodong He, Alex Smola, and Eduard Hovy.
\newblock Hierarchical attention networks for document classification.
\newblock In \emph{Proceedings of the 2016 Conference of the North American
  Chapter of the Association for Computational Linguistics: Human Language
  Technologies}, pages 1480--1489, 2016.

\bibitem[Lin et~al.(2017)Lin, Feng, Santos, Yu, Xiang, Zhou, and
  Bengio]{lin2017structured}
Zhouhan Lin, Minwei Feng, Cicero Nogueira~dos Santos, Mo~Yu, Bing Xiang, Bowen
  Zhou, and Yoshua Bengio.
\newblock A structured self-attentive sentence embedding.
\newblock \emph{arXiv preprint arXiv:1703.03130}, 2017.

\bibitem[Devlin et~al.(2018)Devlin, Chang, Lee, and Toutanova]{devlin2018bert}
Jacob Devlin, Ming-Wei Chang, Kenton Lee, and Kristina Toutanova.
\newblock Bert: Pre-training of deep bidirectional transformers for language
  understanding.
\newblock \emph{arXiv preprint arXiv:1810.04805}, 2018.

\bibitem[Lee et~al.(2019)Lee, Yoon, Kim, Kim, Kim, So, and
  Kang]{lee2019biobert}
Jinhyuk Lee, Wonjin Yoon, Sungdong Kim, Donghyeon Kim, Sunkyu Kim, Chan~Ho So,
  and Jaewoo Kang.
\newblock Biobert: pre-trained biomedical language representation model for
  biomedical text mining.
\newblock \emph{arXiv preprint arXiv:1901.08746}, 2019.

\end{thebibliography}

\newpage
\appendix
\section{Appendix}

\setcounter{table}{0}
\renewcommand{\thetable}{A\arabic{table}}

\begin{table}[!h]
\centering
\caption{A typical example of the PubMed-PICO dataset with sentences and their corresponding annotated labels. The PMID of this abstract is 28449281. }
\label{table:data-examples}
\begin{tabular}{ |>{\raggedright\arraybackslash} m{1.0cm} |>{\raggedright\arraybackslash} m{11.8cm}|}
\hline
 \textbf{Labels} & \textbf{Sentences}                                                                                                                                                                     \\ \hline
 A              & {[}...{]} The aims of the trial were to test for differences between standard 1-and 0.5-mg doses (both twice daily during 8weeks) in (1) abstinence, (2) adherence and (3) side effects.                                               \\ \hline
 M              & Open-label randomized parallel-group controlled trial with 1-year follow-up. [...] Stop-Smoking Clinic of the Virgen Macarena University Hospital in Seville, Spain.                                                        \\ \hline
 P              & The study comprised smokers (n=484), 59.5\% of whom were men with a mean age of 50.67years and a smoking history of 37.5 pack-years.                                                                                                   \\ \hline
 I              & Participants were randomized to 1mg (n=245) versus 0.5mg (n=239) and received behavioural support, which consisted of a baseline visit and six follow-ups during 1year.                                                                \\ \hline
 O              & The primary outcome was continuous self-reported abstinence during 1year, with biochemical verification. {[}...{]} Also measured were baseline demographics, medical history and smoking characteristics.                              \\ \hline
 R              & Abstinence rates at 1year were 46.5\% with 1mg versus 46.4\% with 0.5mg {[}odds ratio (OR)=0.997; 95\% confidence interval (CI) = 0.7-1.43; P=1.0{]}; {[}...{]}                                                                          \\ \hline
 C              & There appears to be no difference in smoking cessation effectiveness between 1mg and 0.5mg varenicline, [...]. \\ \hline
\end{tabular}
\label{table:abstract-example}
\end{table}

\subsection{Sentence Encoder}\label{appendix:sentence-encoder}
\subsubsection{RNN/CNN Based}

As shown in Figure~\ref{figure:model}(B), given a sentence comprising $T$ words, the RNN/CNN based sentence encoder first maps each word to a real-valued vector $\vect{e}$ as its lexical-semantic representation. Word representations are encoded by the column vector in the embedding matrix $W^{\mathit{word}}\in \mathbb{R}^{d^w\times|V|}$, where $d^w$ is the dimension of the word vector and $V$ is the vocabulary of the dataset. Each column $W_i^{\mathit{word}}\in \mathbb{R}^{d^w}$ is the word embedding vector for the $i^{th}$ word in the vocabulary.  The word embeddings $W^{\mathit{word}}$ can be pre-trained on large unlabeled datasets using unsupervised algorithms such as word2vec \citep{mikolov2013distributed}, GloVe \citep{pennington2014glove} and fastText \citep{bojanowski2016enriching}.

Then the sequence of embedding vectors is processed by a bi-directional RNN (bi-RNN) or CNN layer, similar to the ones used in the text classification before \citep{kim2014convolutional}. Whether to use RNN or CNN depends on the dataset size. This layer outputs a sequence of hidden state vectors $\vect{h}_{1:T}$ ($\vect{h}\in \mathbb{R}^{d^{h_s}}$) with each hidden state corresponding to a word. To obtain a single vector to represent the sentence, attentive pooling or self attention \citep{yang2016hierarchical,lin2017structured} is used to aggregate the sequence of hidden state vectors into one. 
Detailed equations are expressed below:
\begin{equation}
\vect{u}_i = \tanh(W_s\vect{h}_i+\vect{b}_s),
\alpha_i=\frac{\exp(\vect{u}_i^\top\vect{u}_s)}{\sum_j \exp(\vect{u}_j^\top\vect{u}_s)},
\vect{s}=\sum_i \alpha_i\vect{h}_i,
\end{equation}
where $\vect{h}_i$ is the hidden state vector of the $i$-th token produced by the sentence encoder, $\vect{u}_s\in \mathbb{R}^{d^{a}}$ is the token level context vector used to measure the relevance or importance of each token with respect to the whole sentence, $W_s$ is a weight matrix, and $\vect{b}_s$ is a bias vector. 

\begin{equation}
A = \text{softmax}(U_s\tanh(W_sH+\vect{b}_s)),
\end{equation}
\begin{equation}
S=AH^T,
\end{equation}
where $H=\begin{bmatrix} h_1 & h_2 & \cdots & h_T\end{bmatrix}\in \mathbb{R}^{d^{hs}\times T}$, $W_s\in \mathbb{R}^{d^a \times d^{hs}}$ is the transformation matrix for soft alignment, $\vect{b}_s\in \mathbb{R}^{d^a}$ is the bias vector, $U_s\in \mathbb{R}^{r \times d^{a}}$ is the token level context matrix used to measure the relevance or importance of each token with respect to the whole sentence, softmax is performed along the second dimension of its input matrix, and $A\in \mathbb{R}^{r \times T}$ is the attention matrix.

Here each row of $U_s$ is a context vector $\vect{u}_s\in \mathbb{R}^{d^{a}}$ and it is expected to reflect an aspect or component of the semantics of a sentence. To represent the overall semantics of the sentence, we use multiple context vectors to focus on different parts of this sentence. Finally, the sentence encoding vector $\vect{s}\in \mathbb{R}^{rd^{hs}}$ is obtained by reshaping the matrix $S$ into a vector.

\paragraph{\textbf{Adversarial Training}}

Deep learning models always suffer from over-fitting, which calls for the fast development of various regularization methods to combat this issue. Here, we apply adversarial and virtual adversarial training as an effective way to regularize the classifier by adding small perturbations to the embeddings while training, as in \citep{miyato2016adversarial}. 
For this, we first normalize the embeddings so that the embeddings and perturbations are on a similar scale, as shown below:
\begin{equation}
\vect{\bar{e}}_k=\frac{\vect{e}_k-E(\vect{e})}{\sqrt[]{\mathit{Var}(\vect{e})}},
E(\vect{e})=\sum_{j=1}^{|V|} f_j\vect{e}_j, \mathit{Var}(\vect{e})=\sum_{j=1}^{|V|} f_j(\vect{e}_j-E(\vect{e}))^2,
\end{equation}
where $f_i$ is the frequency of the $i$-th word based on the statistics of training samples.

We denote the concatenation of a sequence of word embedding vectors $[\vect{\bar{e}}^{(1)},\vect{\bar{e}}^{(2)},...,\vect{\bar{e}}^{(T)}]$ as $\vect{\bar{s}}$ (this sequence can be a sentence or paragraph), and the model conditional probability of the gold label $y$ on $\vect{\bar{s}}$ as $p(y|\vect{\bar{s}};\vect{\theta})$ given the current model parameters $\vect{\theta}$. Then the adversarial perturbation $\vect{r}_{\mathit{adv}}$ is calculated using the following equation:
\begin{equation}
\vect{r}_{\mathit{adv}}=-\epsilon_1 \frac{\vect{g}}{\|\vect{g}\|_2}\ \mathrm{where}\ \vect{g}=\nabla_{\vect{\bar{s}}} \log p(y|\vect{\bar{s}};\vect{\theta}),
\end{equation}
where $\epsilon_1$ controls the scale of the $l_2$-norm of the perturbation. To make the classifier robust to adversarial perturbation, we add the adversarial loss to the original classification loss, which is defined by:
\begin{equation}
L_{\mathit{adv}}(\vect{\theta})=-\frac{1}{N}\sum_{n=1}^N \log p(y_n|\vect{\bar{s}}_n+\vect{r}_{\mathit{adv},n};\vect{\theta}),
\end{equation}
where $N$ is the number of labeled samples. 

For virtual adversarial training, we calculate the following approximated virtual adversarial perturbation:
\begin{equation}
\vect{r}_{\vadv}=\epsilon_2 \frac{\vect{g}}{\|\vect{g}\|_2}\ \mathrm{where}\ \vect{g}=\nabla_{\vect{\bar{s}}+\vect{d}} \mathrm{KL}\big[p(\cdot|\vect{\bar{s}};\vect{\theta})\|p(\cdot|\vect{\bar{s}}+\vect{d};\vect{\theta}) \big],
\end{equation}
where $\vect{d}$ is a small random vector, and $\mathrm{KL}[p\|q]$ stands for the KL divergence between probability distributions $p$ and $q$. Then the virtual adversarial loss is defined as:
\begin{equation}
L_{\vadv}(\vect{\theta})=-\frac{1}{N^{\prime}}\sum_{n=1}^{N^{\prime}} \mathrm{KL}\big[p(\cdot|\vect{\bar{s}}_n;\vect{\theta})\|p(\cdot|\vect{\bar{s}}_n+\vect{r}_{\vadv,n};\vect{\theta}) \big],
\end{equation}
where $N^{\prime}$ is the number of both labeled and unlabeled samples since labels are not needed to calculate the virtual adversarial loss.

When applying adversarial training, the overall loss function is defined as:
\begin{equation}
    L(\vect{\theta})=L_{CE}(\vect{\theta})+\lambda_1L_{\mathit{adv}}(\vect{\theta})+\lambda_2L_{\vadv}(\vect{\theta}),
\end{equation}
where $\lambda_1$ and $\lambda_2$ are the coefficients to scale the contributions of adversarial training losses.

\subsubsection{Transformer Based}

Another sentence encoder we investigate is based on the transformer architecture \citep{vaswani2017attention}, which is illustrated in Figure~\ref{figure:model}(C). Its success partially owes to its long-sequence processing capability, where the hidden state vector for any token contains the context information of any other token in the sequence even though the distance between the two may be very long.  The other important reason is its combined use with inductive transfer learning, which first pre-trains a model over large datasets in a supervised or unsupervised way and then fine-tunes it on the targeted dataset. The induction of knowledge from the large datasets to the small targeted dataset can help generalize the deep models. 

Specifically, in this work, we pre-train a masked language model, named BERT \citep{devlin2018bert}, whose model architecture is a multi-layer bidirectional Transformer encoder \citep{vaswani2017attention} as shown in Figure~\ref{figure:model}(C). BERT was first pre-trained on English Wikipedia and BooksCorpus which contain around 3.3 Billion tokens and then over a large corpus combining PubMed abstracts and PubMed Central (PMC) full-text articles consisting of around 18 billion tokens~\citep{lee2019biobert}. The pre-training and fine-tuning are implemented without supervision, by combining two tasks: one is to predict randomly masked words in a sequence from a bidirectional language model, and the other is a binarized next-sentence prediction task where in two selected text sentences, the model is made to predict whether the latter one is the next sentence to the former one. 

We use this pre-trained BERT as a sentence encoder and extract the output high-dimensional vector corresponding to the first token as the representation vector for this sequence. The parameters of the BERT sentence encoder are updated together with the other parts of the whole model during the training phase.

\begin{table}[!htbp]
\begin{minipage}[t]{.48\linewidth}
\vspace{-3mm}
\caption{Data statistics. ``No.'' indicates the number of sentences; ``C.K.'' means Cohen's Kappa coefficient.}
\label{table:data-stats}
\resizebox{0.95\columnwidth}{!}{\begin{tabular}{lr|lrr}
\toprule
\textbf{PubMed-PICO}  & \textbf{No.} & \textbf{PIBOSO} & \textbf{No.} & \textbf{C.K.} \\ \hline
Aim          & 39073         & Background   & 2557          & 0.70          \\ 
Population   & 27695         & Population   & 812           & 0.63          \\ 
Intervention & 24602         & Intervention & 690           & 0.61          \\ 
Outcome      & 32525         & Outcome      & 2240          & 0.71          \\ 
Method       & 57754         & Study Design & 233           & 0.41          \\ 
Results      & 94133         & Other        & 3396          & 0.67          \\ 
Conclusion   & 44186         &      -        &    -           &  -             \\ 
\bottomrule
\end{tabular}}
\end{minipage}
\qquad
\begin{minipage}[t]{.48\linewidth}
\vspace{-3mm}
\caption{Key hyperparameters for the RNN/CNN based sentence encoder architecture.} \label{table:hyperparameters}
\resizebox{0.95\columnwidth}{!}{\begin{tabular}{lll}
\toprule
\textbf{Hyperparameter}   & \textbf{PubMed-PICO} & \textbf{PIBOSO} \\ \hline
sentence encoder & RNN         & CNN          \\
CNN filter sizes & --          & 2,3,4        \\
$d^{h_s}$                & 100         & 150          \\
$d^{a}$   & 400         & 50           \\
$d^{h_d}$                & 100         & 200          \\
dropout          & 0.2         & 0.4          \\
$\epsilon_1$    & 8           & 4            \\
$\epsilon_2$    & 4           & 4            \\
$\lambda_1$     & 0.2         & 0.3          \\
$\lambda_2$     & 0.05        & 0.3         \\
\bottomrule
\end{tabular}}
\end{minipage}
\end{table}

\subsection{Discussion}

\subsubsection{BERT Pre-training Corpora}

The original BERT sentence encoder was pre-trained using the corpora in the general domain including Wikipedia and BooksCorpus \citep{devlin2018bert}. In general, the semantic and syntactic closeness between the pre-training corpora and target datasets can affect the fine-tuning performance. Therefore we extended the pre-training corpora to include the PubMed abstracts and PMC full text articles, the source of the two datasets used in this work. Table \ref{table:biobert-comparison} compares the PICO element detection performance before and after extending the pre-training corpora. In this table, BERT model means the BERT sentence encoder is pre-trained only over Wikipedia and BooksCorpus, while BioBERT means the pre-trained corpus is the combination of the aforementioned ones and PubMed abstracts and full articles. As can be seen, the improvements brought by closer pre-training corpora exist for all labels, and when the dataset is small, for example the NICTA-PIBOSO, the increment of F1 score is more dramatic. Most strikingly, for instance, BioBERT almost doubles the F1 score for the ``I'' element in the NICTA-PIBOSO dataset compared with BERT. 

\begin{table}[!htbp]
\centering
\small
\vspace{-3mm}
\caption{Comparison of PICO element detection performance. F1 score is reported after 10 folds cross-validation for both datasets. P, I, O, A, M, R, C, B, and SD represent population, intervention, outcome, aim, method, result, conclusion, background, and study design labels, respectively. Dataset PubMed is PubMed-PICO and NICTA is NICTA-PIBOSO.}
\label{table:biobert-comparison}
\resizebox{0.9\columnwidth}{!}{\begin{tabular}{llccccccc}
\hline
\textbf{Dataset} & \textbf{Model}   & \textbf{P (\%)}    & \textbf{I (\%)}    & \textbf{O (\%)}   & \textbf{A (\%)}   & \textbf{M (\%)}  & \textbf{R (\%)} & \textbf{C (\%)} \\ \hline
\multirow{2}{*}{PubMed} & BERT    & 90.2  & 83.7  & 88.0  & 98.7  &  90.4  &  97.3  & 96.7  \\ 
& BioBERT & \textbf{91.0}  & \textbf{84.6}  & \textbf{88.9}  & \textbf{98.8}  & \textbf{91.3}   &  \textbf{97.4}  &  \textbf{97.0} \\ \hline
\textbf{Dataset} & \textbf{Model}   & \textbf{P (\%)}    & \textbf{I (\%)}    & \textbf{O (\%)}   & \textbf{B (\%)}   & \textbf{SD (\%)}  & \textbf{Other (\%)} \\  \hline
\multirow{2}{*}{NICTA} & BERT    & 72.9  & 27.8  & 90.8  & 83.9  &  62.1  &  69.8  &   \\ 
& BioBERT & \textbf{75.4} & \textbf{56.6} & \textbf{91.0} & 83.9 & \textbf{77.1} & \textbf{72.1}   \\ \hline
\end{tabular}}
\end{table}

\subsubsection{Error Analysis}

In order to gain a deeper understanding of the errors made by our method, we have compiled corresponding confusion matrices for the BERT based sentence encoder on the test set of both datasets, which are summarized in Tables \ref{table:biobert-confusion-pico} and \ref{table:biobert-confusion-nicta}. 

\begin{table}[!htbp]
\vspace{-3mm}
\caption{Confusion Matrix.}
\begin{subtable}[t]{.48\linewidth}
\small
\centering
\vspace{-1mm}
\caption{Confusion matrix for the PubMed-PICO dataset. A, M, P, I, O, R and C represent aim, methods, population, intervention, outcome, results, and conclusion. Rows are true labels while columns are predicted labels.} \label{table:biobert-confusion-pico}
\resizebox{0.98\columnwidth}{!}{\begin{tabular}{l|ccccccc}
\toprule
             & \textbf{A}  & \textbf{M} & \textbf{P} & \textbf{I} & \textbf{O} & \textbf{R} & \textbf{C} \\ \hline
\textbf{A}          & 3879 & 22     & 0          & 5            & 11      & 1       & 0          \\
\textbf{M}       & 16   & 5185   & 109        & 263          & 192     & 15      & 0          \\
\textbf{P}   & 13   & 130    & 2456       & 85           & 33      & 34      & 3          \\
\textbf{I} & 7    & 203    & 47         & 2096         & 94      & 11      & 1          \\
\textbf{O}      & 4    & 121    & 19         & 55           & 2921    & 85      & 13         \\
\textbf{R}      & 0    & 14     & 15         & 6            & 52      & 9344    & 99         \\
\textbf{C}   & 0    & 0      & 0          & 0            & 4       & 127    & 4315     \\ \bottomrule
\end{tabular}}
\end{subtable}
\qquad
\begin{subtable}[t]{.48\linewidth}
\centering
\small
\vspace{-1mm}
\caption{Confusion matrix for the NICTA-PIBOSO dataset. B, P, I, O, S, and Ot represent background, population, intervention, outcome, study, and other.} \label{table:biobert-confusion-nicta}
\resizebox{0.98\columnwidth}{!}{\begin{tabular}{l|cccccc}
\toprule
 & \textbf{B} & \textbf{P} & \textbf{I} & \textbf{O} & \textbf{S} & \textbf{Ot} \\ \hline
\textbf{B}   & 440        & 3          & 1            & 20      & 1     & 25    \\
\textbf{P}   & 6          & 48         & 0            & 1       & 1     & 13    \\
\textbf{I} & 4          & 1          & 31           & 5       & 0     & 20    \\
\textbf{O}      & 66         & 5          & 4            & 842     & 0     & 31    \\
\textbf{S}        & 3          & 0          & 0            & 0       & 17    & 1     \\
\textbf{Ot}        & 34         & 12         & 10           & 45      & 2     & 200  \\ \bottomrule
\end{tabular}}
\vspace{-5mm}
\end{subtable}

\end{table}

\begin{table}[!htbp]
\small
\centering
\caption{Error examples for the PubMed-PICO dataset.} 
\label{table:error-pubmed}
{\begin{tabular}{p{1.6cm}p{1.5cm}p{1.5cm}p{7.7cm}}
\hline
\textbf{PubMed ID}                 & \textbf{True Label}   & \textbf{Prediction}   & \textbf{Sentence}                                                                                                                \\ \hline
\multirow{2}{*}{28592575} & Intervention & Outcome      & Anthropomorphic data and blood parameters were collected from our electronic chart programme.                           \\ \cline{2-4} 
                          & Intervention & Outcome      & Water Cr and V data were obtained from the Ontario Water (Stream) Quality Monitoring Network .                          \\ \hline
\multirow{3}{*}{28831814} & Population   & Intervention & Early intervention: Cool Little Kids parenting group programme was implemented.                                         \\ \cline{2-4} 
                          & Population   & Intervention & Primary outcomes: the primary outcomes were child DSM-IV anxiety disorders (assessor blind) and internalising problems. \\ \cline{2-4} 
                          & Outcome      & Intervention & The secondary outcomes were parenting practices and parent mental health.                                               \\ \hline
\end{tabular}}
\end{table}

\begin{table}[t]
\centering
\caption{Error examples for the NICTA-PIBOSO dataset.} \label{table:error-nicta}
{\begin{tabular}{p{2.cm}p{1.5cm}p{9.cm}}
\hline
\textbf{True Label}   & \textbf{Prediction}   & \textbf{Sentence}                                                                                                                             \\ \hline
Other        & Intervention & A medical, surgical, and rehabilitation ward were each randomly assigned to each arm.                                                \\ 
Intervention & Other        & Two oligonucleotides probes for each CYP1A1 polymorphic site were designed and labeled with digoxigenin.                             \\ \hline
Other        & Outcome      & At one month's follow-up, 42 patients had maintained sinus rhythm (group A), and 20 had relapsed into atrial fibrillation (group B). \\ 
Outcome      & Other        & Internal consistency and convergent and divergent validity were established .                                                        \\ \hline
Population   & Other        & Subjects were 24 hemiparetics within 13 months of a stroke , unselected for contracture or spasticity.                               \\ 
Other        & Population   & One hundred and forty-four were followed up at approximately six months, 83 at 12 months, and 71 at 24 months posttrauma.         \\ \hline
\end{tabular}}
\end{table}

By looking at Table \ref{table:biobert-confusion-pico}, we find that the most confusion for all P/I/O labels in the PubMed-PICO dataset comes from the ``Method'' label, which makes sense because in terms of a broader rhetorical structure of an abstract, P/I/O elements can all be categorized into the ``Method'' group. In other words, P/I/O elements can be deemed as specific subsections of an upper ``Method'' section. The second major confusion sources for P, I and O elements are I, O and Results, respectively. These confusions are also reasonable since there are some cases where these labels are ambiguous. For example, as shown in Table \ref{table:error-pubmed}, two sentences from the PubMed abstract ``28592575'' belong to the ``intervention'' label as ground truth but are predicted as ``outcome''. By looking at these two sentences, it is indeed ambiguous which label they should be categorized into. There are also two other smaller error sources: 1.~In some cases, the ground truth is erroneous; 2.~The CRF module at the top of our model would cause some friction to the transition between different subsequent labels, that is, the CRF module prefers to outputting the same label as the former one for two subsequent predictions. These two scenarios are exemplified by the PubMed abstract ``28831814'' in Table \ref{table:error-pubmed}. The sentence ``Early intervention: Cool Little Kids parenting group programme was implemented.'' should belong to the ``intervention'' label but the ground truth is ``population'', which is because the ground truth is compiled from the section headings and when the authors were writing the structured abstracts, some errors could happen for the headings. On the other hand, the following two sentences in the same abstract should be ``outcomes'', but the CRF module prefers the model to output the same labels as the previous label, which is ``intervention''. 

For the confusion matrix of NICTA-PIBOSO dataset at Table \ref{table:biobert-confusion-nicta}, we observe that the major confusion source for all labels is the ``other'' label, which emerges when the sentence cannot be categorized into any other labels. Such a miscellaneous label could potentially carry some properties of all other labels, thus leading to much confusion, which is exemplified by Table \ref{table:error-nicta}. From this table, we see that there is indeed much confusion between the ``other'' label and P/I/O elements. 


\end{document}